\documentclass[10pt,twocolumn,letterpaper]{article}

\usepackage{wacv}              

\usepackage{graphicx}
\usepackage{amsmath}
\usepackage{amssymb}
\usepackage{booktabs}
\usepackage{multirow}
\usepackage{float}

\usepackage[pagebackref,breaklinks,colorlinks]{hyperref}

\usepackage[capitalize]{cleveref}
\crefname{section}{Sec.}{Secs.}
\Crefname{section}{Section}{Sections}
\Crefname{table}{Table}{Tables}
\crefname{table}{Tab.}{Tabs.}

\def\wacvPaperID{910} 
\def\confName{WACV}
\def\confYear{2025}

\begin{document}
%
\thispagestyle{empty}
\onecolumn
\linespread{1.2}\selectfont{}
{\noindent\Huge IEEE Copyright Notice}\\[1pt]

{\noindent\large Copyright (c) 2023 IEEE

\noindent Personal use of this material is permitted. Permission from IEEE must be obtained for all other uses, in any current or future media, including reprinting/republishing this material for advertising or promotional purposes, creating new collective works, for resale or redistribution to servers or lists, or reuse of any copyrighted component of this work in other works.}\\[1em]

{\noindent\Large Accepted to be published in: {IEEE/CVF} Winter Conference on Applications of Computer Vision (WACV'25), February 28 -- March 4, 2025.}\\[1in]

{\noindent\large Cite as:}\\[1pt]

{\setlength{\fboxrule}{1pt}
 \fbox{\parbox{0.65\textwidth}{A. Sacilotti, S. F. Santos, N. Sebe, and J. Almeida, ``Transferable-guided Attention Is All You Need for Video Domain Adaptation'' in \emph{{IEEE/CVF} Winter Conference on Applications of Computer Vision (WACV)}, Tucson, AZ, USA, 2025, pp. 1--11}}}\\[1in] 
 
{\noindent\large BibTeX:}\\[1pt]

{\setlength{\fboxrule}{1pt}
 \fbox{\parbox{0.95\textwidth}{
 @InProceedings\{WACV\_2025\_Sacilotti,
 
 \begin{tabular}{lll}
  & author    & = \{A. \{Sacilotti\} and 
                    S. F. \{Santos\} and
                    N. \{Sebe\} and
                    J. \{Almeida\}\},\\
			   
  & title     & = \{Transferable-guided Attention Is All You Need for Video Domain Adaptation\}, \\
			   
  & pages     & = \{1--11\},\\
  
  & booktitle & = \{{IEEE/CVF} Winter Conference on Applications of Computer Vision (WACV)\},\\
  
  & address   & = \{Tucson, AZ, USA\},\\
  
  & month     & = \{February 28 -- March 4\},\\
  
  & year      & = \{2025\},\\
  
  & publisher & = \{\{IEEE\}\},\\
  
  \end{tabular}
  
\}
 }}}

\twocolumn
\linespread{1}\selectfont{}
\clearpage

\title{Transferable-guided Attention Is All You Need for Video Domain Adaptation}

\author{%
André Sacilotti$^1$ 
\hskip1em
Samuel Felipe dos Santos$^2$
\hskip1em
Nicu Sebe$^3$
\hskip1em
Jurandy Almeida$^2$ \\[1em]
\begin{tabular}{ccc}
$^1$University of São Paulo &
$^2$Federal University of São Carlos &
$^3$University of Trento\\
{\tt\small andre.sacilotti@usp.br} &
{\tt\small \{felipe.samuel,jurandy.almeida\}@ufscar.br} &
{\tt\small niculae.sebe@unitn.it} \\
\end{tabular}}
\maketitle

\begin{abstract}
   Unsupervised domain adaptation~(UDA) in videos is a challenging task that remains not well explored compared to image-based UDA techniques.
    Although vision transformers~(ViT) achieve state-of-the-art performance in many computer vision tasks, their use in video UDA has been little explored. 
    Our key idea is to use transformer layers as a feature encoder and incorporate spatial and temporal transferability relationships into the attention mechanism. 
    A Transferable-guided Attention~(TransferAttn) framework is then developed to exploit the capacity of the transformer to adapt cross-domain knowledge across different backbones. To improve the transferability of ViT, we introduce a novel and effective module, named Domain Transferable-guided Attention Block~(DTAB). DTAB compels ViT to focus on the spatio-temporal transferability relationship among video frames by changing the self-attention mechanism to a transferability attention mechanism.
    Extensive experiments were conducted on UCF-HMDB, Kinetics-Gameplay, and Kinetics-NEC Drone datasets, with different backbones, like ResNet101, I3D, and STAM, to verify the effectiveness of TransferAttn compared with state-of-the-art approaches. Also, we demonstrate that DTAB yields performance gains when applied to other state-of-the-art transformer-based UDA methods from both video and image domains. Our code is available at \url{https://github.com/Andre-Sacilotti/transferattn-project-code}.
\end{abstract}


\section{Introduction}
\label{sec:intro}

\begin{figure*}[!htb]
  \centering
  \includegraphics[width=0.76\textwidth]{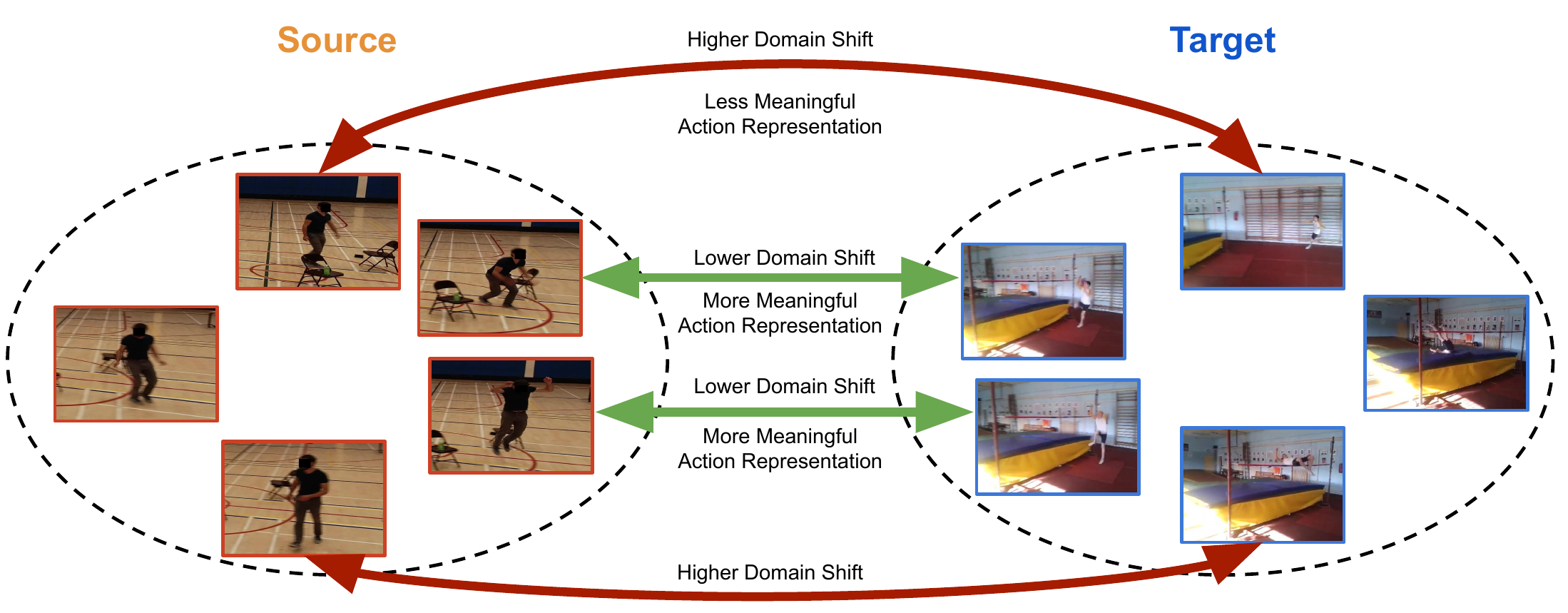}
  \caption{Main intuition behind our method. In this toy example, some frames in the \textit{Jumping} class have more meaningful representation about the action, like the pose about to start the jump movement. Also, due to the capability to represent the important steps of the action, such frames can have minor domain shift compared to the rest of the video.}
  \label{fig:toyexample1}
\end{figure*}

With the popularization of social media platforms focused on user-generated content, a huge volume of data is generated, for instance, 720,000 of hours of video content is uploaded to YouTube daily\footnote{\url{https://www.demandsage.com/youtube-stats/}}.
The cataloging and searching of this content is necessary, however, manually analyzing this immense amount of content is practically impossible, making video analysis tasks crucial.

Among the several video analysis tasks, action recognition is one of the most popular and challenging ones since there is a significant number of variations in the manner the action can be carried out and captured, for example, speed, duration, camera, and actor movement, occlusion, etc~\cite{udavt}.

Various deep learning methods for action recognition are available in the literature. These approaches can be classified based on how they handle the temporal dimension. Some use 3D models to capture spatial and temporal features, while others treat spatial and temporal data separately or employ Recurrent Neural Networks~(RNNs) to model the temporal dynamics~\cite{kong2022human}. 
Despite all the advances, the temporal structure of videos still poses some challenges for training deep learning models~\cite{CVPR_2018_Huang}.
Human costs are high, as many video annotations are needed to yield good results. Obtaining and annotating a desirable amount of data is difficult for many application domains, requiring significant human effort and specific knowledge~\cite{7508942}.

Unsupervised Domain Adaptation~(UDA) can be used to reduce the cost of manually annotating data.
In these strategies, the model is trained with labeled data from a source domain and unlabeled data from a target domain to perform well on the target domain's test set.
Since there is a domain change between source and target, UDA methods must deal with the distribution mismatch generated by the domain gap, since the domains might have different backgrounds, illumination, camera position, etc~\cite{ma2ltd}. 
Several methods have been proposed to address this issue, e.g., adversarial-based methods~\cite{dann, 8099799, 10.5555/3045118.3045244}, metric-based methods~\cite{10.1007/978-3-319-13560-1_76,10.5555/3305890.3305909}, and more recently, transformer-based methods~\cite{DBLP:journals/corr/abs-2109-06165,10030518}, achieving remarkable results.
However, most of these works are for image UDA, and video UDA is considerably less explored and significantly more challenging, as it requires handling the temporal aspects of the data~\cite{udavt}.

Only a few recent works~\cite{mm-sada, ta3n, mix-dann, transvae, co2a, cleanadapt, ma2ltd, 10.1145/3503161.3548009, udavt, 10204931} tackle video UDA for action recognition using deep learning with strategies like contrastive learning, cross-domain attention mechanisms, self-supervised learning, and multi-modalities of data.
An amount even lower of works~\cite{udavt, 10.1145/3503161.3548009} explore Vision Transformer~(ViT) architectures.

Although existing methods have improved the performance of UDA for action recognition, some limitations remains unexplored. First, they focus on aligning frames with a larger spatial and temporal domain gap across an entire video, but, as shown in Figure~\ref{fig:toyexample1}, we hypothesize that frames with a smaller domain gap have a more meaningful action representation and can improve the adaptability. Also, none of them investigate how to improve ViT for video UDA.

To address the mentioned issues, we propose a novel method for video UDA, called Transferable-guided Attention (TransferAttn), which improves the potential of transformer architecture.
Our method uses pre-trained frozen backbones to extract frame-by-frame features from videos.
A transformer encoder is used to reduce the domain gap and learn temporal relationships among frames.
The encoder also includes our proposed transformer block, named Domain Transferable-guided Attention Block~(DTAB), which introduces a new attention mechanism.

We evaluate our approach on three well-known video UDA benchmarks for action
recognition,  namely UCF $\leftrightarrow$ HDMB$_{full}$~\cite{ta3n}, Kinetics $\rightarrow$ Gameplay~\cite{ta3n}, and Kinetics $\rightarrow$ NEC-Drone~\cite{necdrone}, where we outperform the other state-of-the-art methods. 
We also integrated our proposed DTAB module into other state-of-the-art transformer architectures for UDA, showing that it may increase their performance.

The main contributions of this paper are the following:
\begin{itemize}\itemsep=1pt
    \item To the best of our knowledge, we are the first to present a backbone-independent transformer architecture for video UDA. Our experiments showed the effectiveness of the transformer encoder in extracting fine-grained spatio-temporal transferable representations.
    \item We propose DTAB, a novel transferable transformer block for UDA. Our method employs a new attention mechanism that improves adaptation and domain transferability. Also, we show the positive effect of applying the DTAB module to other state-of-the-art UDA methods, both for videos and images.
    \item We conduct extensive experiments on several benchmarks, setting a new state-of-the-art result in three different cross-domain datasets. Also, our ablation study shows the positive effect of each part of our approach.
\end{itemize}

\section{Related Work}
\label{sec:related}

\subsection{Video-based Action Recognition} 
Action recognition methods have been extensively studied with the advent of deep learning, especially with the introduction of large-scale video datasets, such as Kinetics, Moments-In-Time, YouTube Sports 1M, and Youtube 8M~\cite{6165309}. BEAR~\cite{deng2023BEAR} states a new benchmark in action recognition, which is made to cover a diverse set of real-world applications. Deep learning-based solutions can be divided into three categories according to how they model the temporal dimension~\cite{kong2022human}: (1) space-time networks, (2) multi-stream networks, and (3) hybrid models. Space-time networks use 3D convolutions to maintain temporal information, inflating 2D kernels to 3D, like C3D~\cite{7410867} and I3D~\cite{i3d}.
Multi-stream networks employ different models to deal with spatial (usually RGB images) and temporal (usually optical flow) information, like TSN~\cite{wang2018temporal}, which applies temporal sampling; and TDN~\cite{wang2021tdn}, which has modules to capture short-term and long-term motion (across segments). 
Hybrid models integrate recurrent networks, like LSTMs~\cite{donahue2015long,yue2015beyond,wu2015modeling} and Temporal CNNs~\cite{ke2017new}, on top of CNNs. Skeleton data, like body joint information, can also be utilized~\cite{shahroudy2016ntu,zhu2016co,liu2016spatio,ke2017new} and recent works~\cite{yan2018spatial,si2019attention} show that graph convolution obtains superior performance to RNNs and Temporal CNNs on capturing information from joints~\cite{kong2022human}.  Kim~et~al.~\cite{kim2024adversarial} presents a novelty training approach to make models robust to distribution shifts.

\subsection{Unsupervised Domain Adaptation} 
Unsupervised domain adaptation~(UDA) in the image domain has a wide range of strategies to address the domain shift. A standard option is the adversarial-based methods~\cite{dann, 8099799, 10.5555/3045118.3045244, lai2024empowering}, which use a domain discriminator while maximizing the feature extractor loss through a min-max optimization game, similar to Generative Adversarial Networks~(GAN)~\cite{NIPS2014_5ca3e9b1} training, minimizing the domain gap. In addition, the metric-based methods aim to reduce the domain gap by learning domain-invariant features through discrepancy metrics, like Maximum Mean Discrepancy (MMD)~\cite{10.1007/978-3-319-13560-1_76} and Joint Adaptation Networks~(JAN)~\cite{10.5555/3305890.3305909}, that incorporate a loss metric computing the discrepancy between the domain features and aim to reduce that metric to minimize the domain shift. Driven by the success of ViTs, CDTrans~\cite{DBLP:journals/corr/abs-2109-06165} adopts a three-branch cross transformer that proves to be noisy-robust. On the other side, TVT~\cite{10030518} employs a transferability metric as a weight into class token attention weight. Although TVT~\cite{10030518} shows great results injecting transferability into the class token weight, it lacks two essential points: i) TVT~\cite{10030518} does not use spatial relation transferability; ii) As an image UDA, it does not incorporate the temporal relation transferability.

\subsection{UDA for Action Recognition} 
Although there are several possible applications of UDA for action recognition in real-world problems, only a limited number of recent studies have tackled this challenging task~\cite{mm-sada, ta3n, mix-dann, transvae, co2a, cleanadapt, ma2ltd, 10204931}. TA$^3$N~\cite{ta3n} proposes a domain attention mechanism that focuses on the temporal dynamics of the videos. MA$^2$LT-D~\cite{ma2ltd} generates multi-level temporal features with multiple domain discriminators.  Level-wise attention weights are calculated by domain confusion and features are aggregated by attention determined by the domain discriminators. Other approaches use multiple modalities of data, like MM-SADA~\cite{mm-sada}, where self-supervision among modalities is used, and MixDANN~\cite{mix-dann}, which dynamically estimates the most adaptable modality and uses it as a teacher to the others. CleanAdapt~\cite{cleanadapt} tackles the source-free video domain adaptation problem using a model pre-trained on the source domain to generate noisy labels for the target domain, and then the likely correct ones are used to fine-tune the model. STHC~\cite{10204931} addresses the source-free domain using spatial and temporal augmentation. In a different direction, TranSVAE~\cite{transvae} handles spatial and temporal domain divergence separately by constraining different sets of latent factors.

Although transformers can obtain state-of-the-art performance, only a few works for video UDA exist. UDAVT~\cite{udavt} is a recent work that leverages the STAM visual transformer~\cite{stam} and proposes a domain alignment loss based on the Information Bottleneck~(IB) principle to learn domain invariant features. Also, MTRAN~\cite{10.1145/3503161.3548009}, which depends on 3D backbones, uses a transformer layer inspired by ViViT~\cite{arnab2021vivit}, where each token is a 16-frame clip representation. Although UDAVT and MTRAN show great results incorporating the transformer mechanism, the UDAVT architecture strictly depends on transformer backbones that deal separately with spatial and temporal relations, like STAM~\cite{stam}. At the same time, MTRAN is dependent on 3D backbones, and the attention relation is done on clip-level pooled features, lacking a more fine-grained frame-level relation. Also, none of them exploit ways to improve knowledge transferring in the transformer mechanism.


\section{Our Approach}
\label{sec:method}

First, in Section~\ref{subsec:baseline}, we introduce our baseline model, which uses ViT as a feature encoder for a simple adversarial domain adaptation approach. Then, in Section~\ref{subsec:transferattn}, we introduce our domain transferable-guided attention block, called DTAB, and its components.


\subsection{Baseline Model}
\label{subsec:baseline}

\begin{figure}[!htb]
  \centering
  \includegraphics[width=0.9\columnwidth]{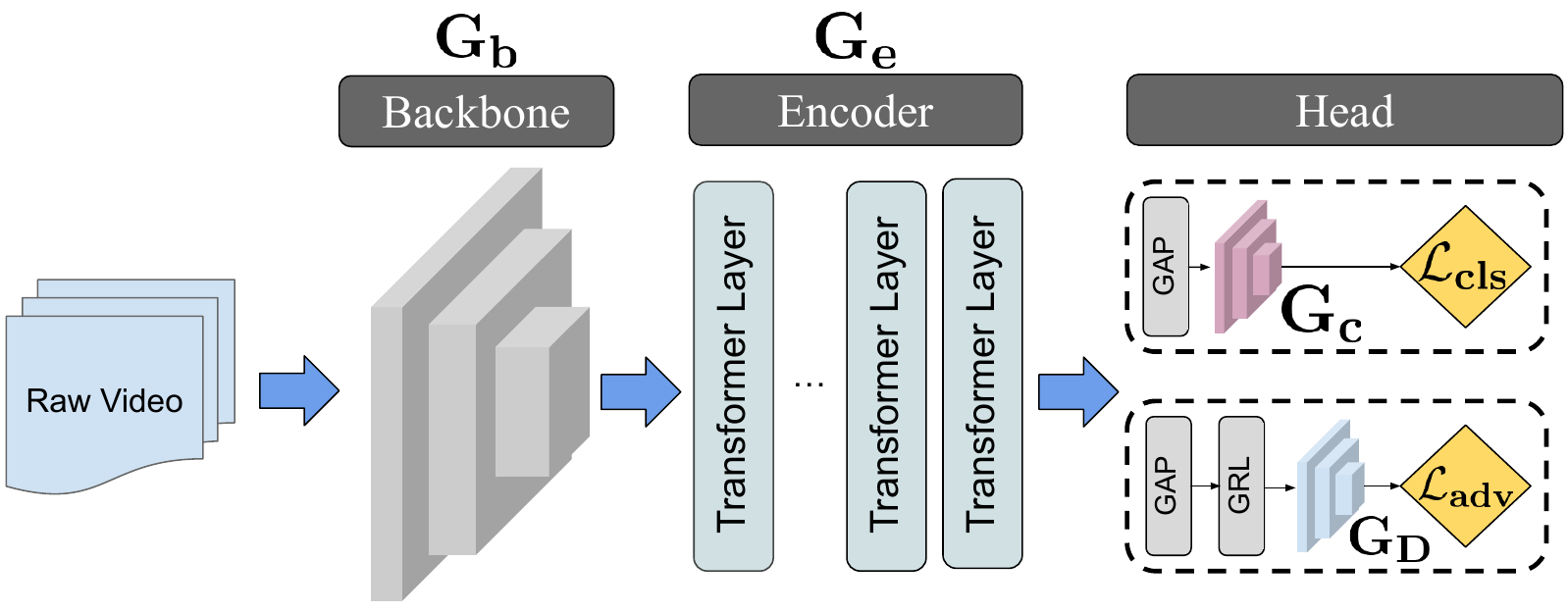}
  \caption{A baseline architecture trained with an adversarial loss $\mathcal{L}_{adv}$ and a classification loss $\mathcal{L}_{cls}$. The backbone $G_b$ extracts frame-level features and the encoder $G_e$ learns meaningful semantic spatio-temporal representations.}
  \label{fig:baseline}
\end{figure}

Given the success of ViTs in diverse computer vision tasks~\cite{liu2021swin, girdhar2019video, carion2020end}, we built a baseline architecture that leverages a ViT encoder to produce better spatio-temporal features for action recognition and an adversarial-based method for UDA, as shown in Figure~\ref{fig:baseline}.

\begin{figure*}[!htb]
  \centering
  \includegraphics[width=0.85\textwidth]{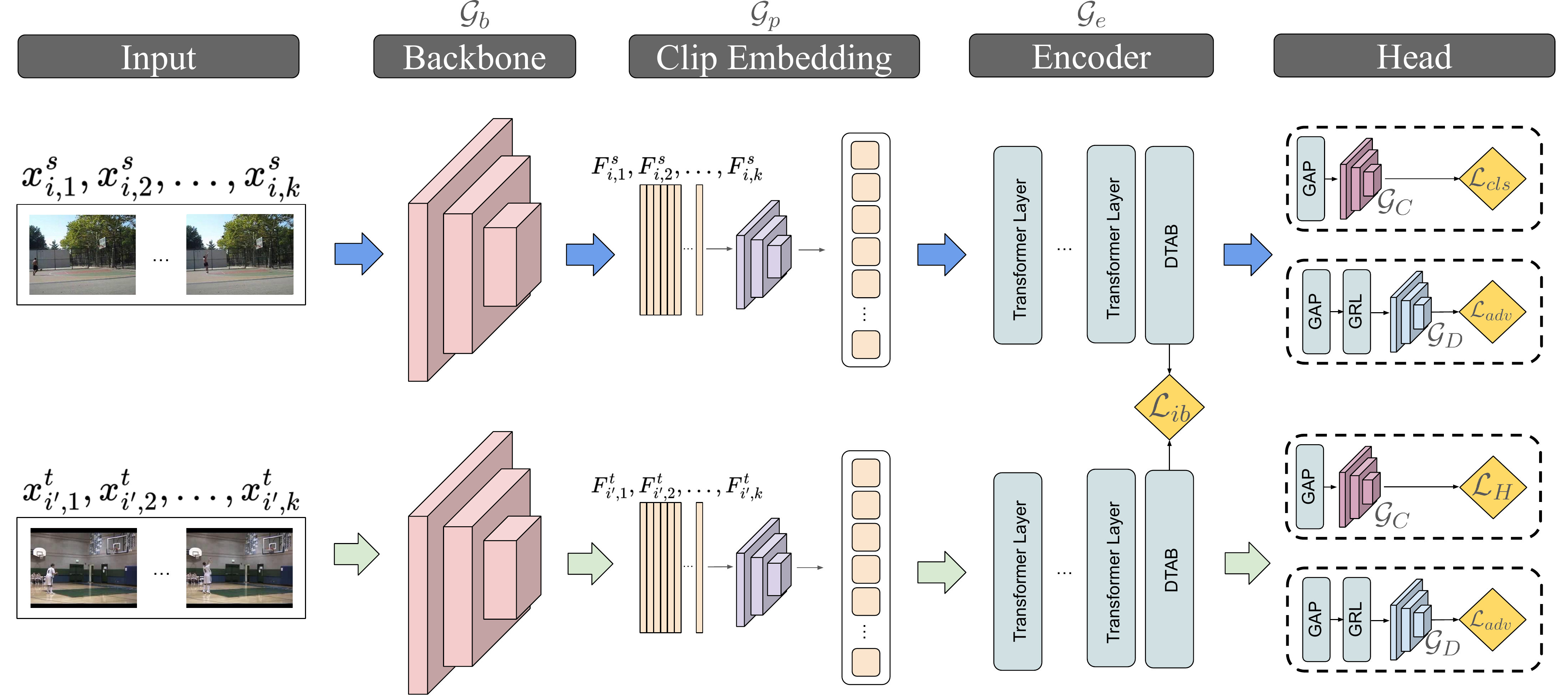}
  \caption{Overview of our TransferAttn. The video frames are fed into a fixed backbone to extract frame-by-frame features, followed by a clip embedding to map frames into tokens. The embeddings are fed into a sequence of transformers to extract transferable spatio-temporal information. The adaptation branch for adversarial domain discrimination uses fine-grained representations from the transformer encoder.}
  \label{fig:entire_model}
\end{figure*}

Overall, the baseline architecture consists of a backbone, an encoder, a classification head, and an adaptation head. The backbone ($G_{b}$) is fixed and not trained. The encoder ($G_{e}$) comprises $L$ transformer layers, with $h$ attention heads and a hidden size of $d$. Before the encoder $G_{e}$, a patch embedding ($G_{p}$) may be used to map the features produced by the backbone $G_{b}$ to the input size of the encoder $G_{e}$. 


Let $\mathcal{S}$ be a labeled source domain sharing the same set $\mathcal{Y}$ of categories with an unlabeled target domain $\mathcal{T}$, where $\mathcal{S} = \{ (x^s_i, y^s_i) \}_{i=1}^{N_s}$ is a set of tuples composed of $N_s$ videos $x^s_i$ and their respective label $y^s_i$. 
Since we do not know the labels of the target domain beforehand, the set $\mathcal{T}=\{x^t_i \}_{i=1}^{N_t}$ comprises the target videos only.
Once $k$ frames are uniformly sampled from each video, we denote the $j$-th frame from the $i$-th video as $x^s_{i,j}$ for the source domain and $x^t_{i,j}$ for the target domain.
For convenience, we refer to the features extracted from the backbone $G_{b}$ for the $i$-th video and the $j$-th frame from the source domain as $F^s_{i,j}$, and $F^t_{i,j}$ for the target domain. 
Also, we denote $f^s_{i}$ and $f^t_{i}$ as a Global Average Pooling~(GAP) over the frame-level features extracted by the encoder $G_{e}$ for the $i$-th video from the source and target domains, respectively.


The classification head contains a classifier ($G_{C}$). 
Unlike previous works, where the classifier $G_{C}$ is trainable, in our framework, the classifier $G_{C}$ is not trained. Therefore, instead of learning a mapping from features to classes, our model uses a MLP classifier with fixed random weights. A similar idea is also adopted by UDAVT~\cite{udavt} in the projection head used to map source and target domain features. By using a fixed classifier, the classification boundaries do no change during training and the encoder $G_{e}$ is enforced to learn a feature space that better fits to the class separability imposed by the classifier $G_{C}$. The motivation for using a fixed classifier is to prevent it from learning a projection that might overfit the source domain data. This way, we make learning class discrimination a responsibility of the encoder $G_{e}$.
%
Although the classification branch is not trained, the classification ($\mathcal{L}_{cls}$) and entropy ($\mathcal{L}_{H}$) losses, as given by Equations~\ref{eq:Lcls}~and~\ref{eq:LH}, respectively, are used to train the model, ensuring that the encoder $G_{e}$ learns a feature space from both source and target domain data. 
As far as we know, we are the first to propose this idea for video UDA.
\begin{equation}
\label{eq:Lcls}
    \mathcal{L}_{cls} = -\frac{1}{N_s}\sum_{i=1}^{N_s} y^s_i \cdot \log 
    G_{C}(f^s_i)
\end{equation}
\begin{equation}
\label{eq:LH}
    \mathcal{L}_{H} = -\frac{1}{N_t}\sum_{i=1}^{N_t} G_{C}(f^t_i) \cdot \log G_{C}(f^t_i)
\end{equation}

The adaptation branch is composed of a Gradient Reversal Layer~(GRL) followed by a domain discriminator ($G_{D}$), which is a MLP classifier trained to identify if a video is from the source or target domain. At the same time, as the gradients are inverted in the GRL, the encoder $G_{e}$ is trained to deceive the domain discriminator $G_{D}$, playing a min-max game. This is achieved with the adversarial domain discriminator loss ($\mathcal{L}_{adv}$), as in Equation~\ref{eq:ladv}.
\begin{equation}
\label{eq:ladv}
\small
    \mathcal{L}_{adv} = -\frac{1}{N_s}\sum_{i=1}^{N_s} \log 
    G_{D}(\mathrm{GRL}(f^s_i)) -\frac{1}{N_t}\sum_{i=1}^{N_t} \log 
    G_{D}(\mathrm{GRL}(f^t_i))
\end{equation}

Therefore, the overall loss used to train the baseline model is a weighted sum of three terms: the classification loss $\mathcal{L}^{s}_{cls}$ (Equation~\ref{eq:Lcls}), the entropy loss $\mathcal{L}^{t}_{H}$ (Equation~\ref{eq:LH}), and the adversarial loss $\mathcal{L}_{adv}$ (Equation~\ref{eq:ladv}).

\begin{figure*}[!htb]
  \centering
  \includegraphics[width=0.92\textwidth]{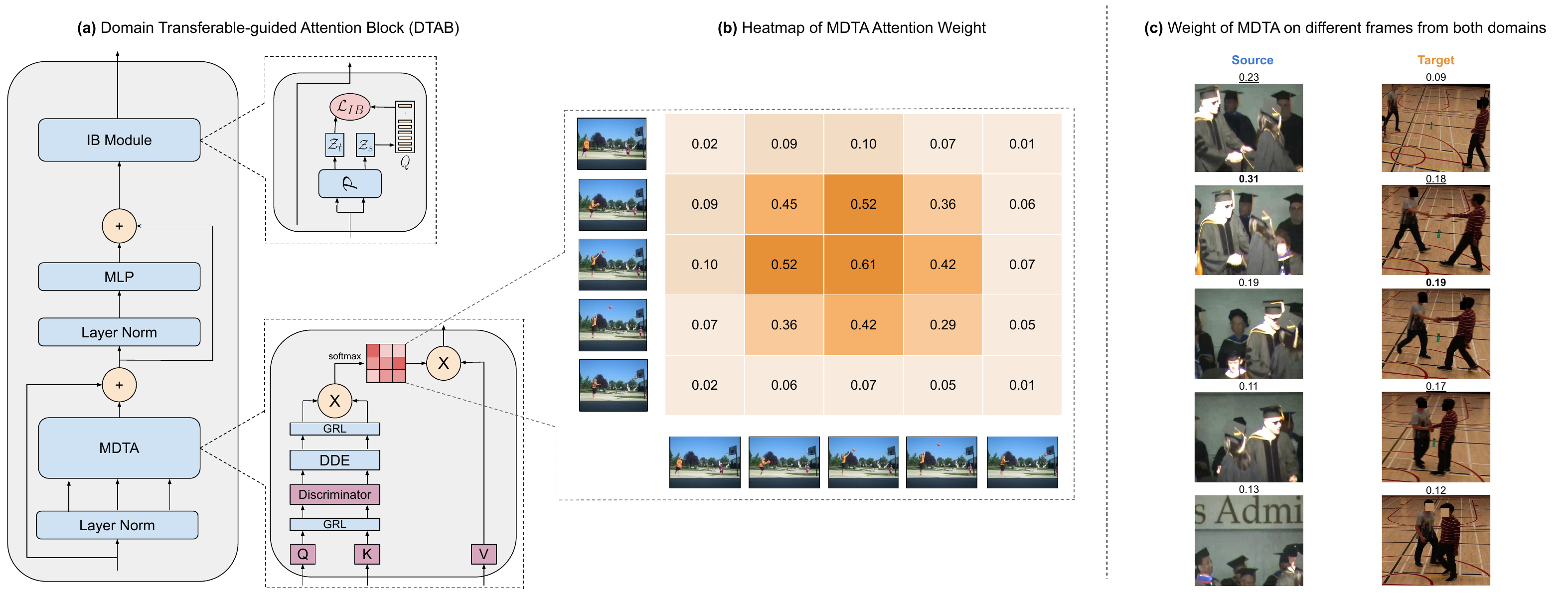}
  \caption{DTAB overview. (a) DTAB follows a standard transformer block, except for our novel MDTA mechanism and the layer-wise IB calculation. (b) Heatmap of transferable-attention weights, showing how MDTA focuses on frames that are more transferable between domains and also brings more meaningful information about the action. (c) Temporal attention visualization compared between domains.
  }
  \label{fig:DTB_block}
\end{figure*}

\subsection{TransferAttn: Transferable-guided Attention}
\label{subsec:transferattn}

The baseline architecture with standard ViT encoder, as shown in Tables~\ref{table:ucf_hmdb_results},~\ref{table:k_g_results},~\ref{table:k_n_results}, achieves limited improvement for video UDA in the action recognition task. Motivated by this observation, we hypothesize that merely using the standard self-attention ViT is insufficient to fully leverage the transferable capabilities of ViT, as it overlooks two crucial factors: (\textit{i}) it does not consider inter-domain specificity and (\textit{ii}) it focus only on intra-domain frames similarity.

To support our hypothesis, we propose TransferAttn, an improvement of our baseline model, as shown in Figure~\ref{fig:entire_model}. The key difference between TransferAttn and our baseline model is in the encoder $G_{e}$. To ensure the encoder $G_{e}$ is capable of learning a better feature space, we design a novel transformer module to facilitate the domain alignment, named Domain Transferable-guided Attention Block~(DTAB), as shown in Figure~\ref{fig:DTB_block}. 
Roughly, DTAB is a transformer block where the self-attention layer is replaced by a new attention mechanism and a domain alignment module is added before the output (see Figure~\ref{fig:DTB_block}(a)).

Instead of using self-attention, which gives attention to similar tokens or patches (in our case, frames) only within a same domain context, we propose the Multi-head Domain Transferable-guided Attention~(MDTA), a transferable-guided attention that acts as a cross-domain attention. This mechanism weights frames based on how difficult it is to predict which domain they belong to (see Figure~\ref{fig:DTB_block}(b)), pushing the model to focus on local temporal features that have a small domain gap. Unlike methods that prioritize frames with larger domain gaps, like TA3N~\cite{ta3n} and MA2LT-D~\cite{ma2ltd}, the reasoning behind MDTA is to highlight portions of a video that are more transferable, or in other words, similar across different domains.
As an example, Figure~\ref{fig:DTB_block}(c) shows the weights produced by MDTA for each frame of a video taken from source and target domains. Note that MDTA enables the model to focus on more significant frames in the action \textit{Hand Shaking} that are more transferable between different domains.

MDTA does not align the source and target domains, but only keeps the attention on portions of a video that are likely to deceive a discriminator in predicting which domain they come from. To perform domain alignment, DTAB exploits the Information Bottleneck~(IB) principle~\cite{2020arXiv200308264K, 2020arXiv200610297P}. The IB mechanism adopted in DTAB was introduced by UDAVT~\cite{udavt}, but unlike UDAVT, DTAB applies IB to local features. The IB mechanism helps to find a shared representation space where the weighted local features from different domains coming out from MDTA are mapped so that the shared information between domains is maximized. DTAB works not only on our TransferAttn, but also improves the results of other state-of-the-art methods, both for video and image UDA, as reported in Tables~\ref{table:udavt_with_our_transformer}~and~\ref{table:tvt_with_our_transformer}. As far we know, we are the first to propose this idea for video UDA.


\textbf{Multi-head Domain Transferable-guided Attention.} 
To overpass the main bottlenecks related to the standard self-attention mechanism, we propose the Multi-head Domain Transferable-guided Attention~(MDTA), an new attention mechanism that leverages spatio-temporal information to give more weight to frames or patches that are more similar across domains and still semantic meaningful.
%
Similar to self-attention, MDTA is computed as in Equations~\ref{eq:mtda}~and~\ref{eq:dta},
\begin{gather}
    \label{eq:mtda}
    \mathrm{MDTA}(Q, K, V) = \mathrm{Concat}(\{\mathrm{DTA}_i\}_{i=1}^{h}) \cdot W_{O}, \\
     \label{eq:dta}
    \mathrm{DTA}_i = \mathrm{softmax}\left(\frac{
    Q^{'}\cdot K^{'T}}{\sqrt{d_h}}\right) \cdot V \cdot W^{V}_{i},
\end{gather}
\noindent where $W^{Q}_{i}$, $W^{K}_{i}$, and $W^{V}_{i}$ are linear projections learned separately for each of the $h$ heads; and $W^{O}$ is the learned linear projection of the concatenation.

Unlike self-attention, in MDTA, the attention scores are computed by taking the dot product between vectors $Q^{'} = \mathrm{GRL}(\mathrm{DDE}(G_{D'}(\mathrm{GRL}(QW^{Q}_{i}))))$ and $K^{'} = \mathrm{GRL}(\mathrm{DDE}(G_{D'}(\mathrm{GRL}(KW^{K}_{i}))))$, where the Domain Discriminator Error~(DDE) is a weighting function given by Equation~\ref{eq:dde} and $G_{D'}$ is a domain discriminator that predicts which domain the frames of a video belong to. The purpose of the double GRL is to force the domain discriminator $G_{D'}$ to learn domain-separable class-invariant features.

\begin{equation}\abovedisplayskip=0pt
    \mathrm{DDE}(x) = \begin{cases}
    \log x, & \text{if $x$ is from source} \\
    \log (1-x), & \text{otherwise}
    \end{cases}
    \label{eq:dde}
\end{equation}

This operation results in a matrix that defines which frames are more or less transferable, as depicted in Figure~\ref{fig:DTB_block}(b), where the higher transferability values are given to the middle frames, probably because they are more meaningful and where the action actually takes place and are therefore likely similar to other domains. In other words, if DDE goes to one, it is more likely to deceive the domain discriminator $G_{D'}$ because its features are not easy to predict which domain they come from and should be more important when classifying the video action. 






\noindent
\textbf{Information Bottleneck Module.} 
Since MDTA does not align the source and target domains, we propose to integrate it with a Information Bottleneck~(IB) module~\cite{udavt}. To perform domain alignment, this module calculates a loss function given by Equation~\ref{eq:ib},
\begin{equation}
\label{eq:ib}
    \mathcal{L}_{ib} = \sum^{m}_{i=1}(1 - C_{i,i})^2 + \lambda \sum^{m}_{i=1} \sum^{m}_{j\ne i}(C_{i,j})^2
\end{equation}

\noindent where $C$ is a cross-correlation matrix whose elements $C_{i,j} = \frac{\sum^{B}_{b} z^s_{i,b} \cdot z^t_{j,b}}{\sqrt{\sum^{B}_{b} (z^s_{i,b})^2}\sqrt{\sum^{B}_{b} (z^t_{j,b})^2}}$ are computed over mean-centered representations ${z}^{s}_{i}$ and ${z}^{t}_{j}$ of the source and target domains, respectively, taken from a batch of $B$ out of a total of $m$ features. 


The loss function used to train our TransferAttn model is a weighted sum of four terms: classification loss ($\mathcal{L}^{s}_{cls}$), entropy loss ($\mathcal{L}^{t}_{H}$), adversarial loss ($\mathcal{L}_{adv}$), and IB loss ($\mathcal{L}_{ib}$).


\section{Experiments and Results}
\label{sec:experiments}

To assess the effectiveness of our approach, we carried out extensive experiments on several datasets for video UDA and compared against state-of-the-art methods.

\subsection{Datasets}
\label{subsec:datasets}

Experiments were conducted on 3 different benchmarks:

\textbf{UCF }$\leftrightarrow$ \textbf{HDMB}$_{full}$~\cite{ta3n}   is one of the most widely used, containing a subset of videos from two public datasets, UCF101~\cite{ucf101} and HMDB51~\cite{hmdb51}, representing a total of 3209 videos and 12 classes.

\textbf{Kinetics }$\rightarrow$ \textbf{Gameplay}~\cite{ta3n} is a non-public dataset that contains a subset of videos from the well-known Kinetics-400~\cite{k400} and a private gameplay dataset~\cite{ta3n}, containing 49998 videos and 30 classes.

\textbf{Kinetics }$\rightarrow$ \textbf{NEC-Drone}~\cite{necdrone} is a public dataset that contains videos from Kinetics-600~\cite{k600} and NEC-Drone. The dataset contains 10118 videos and 7 classes. For a fair comparison, we used the cropped version of NEC-Drone~\cite{udavt}.


\subsection{Experimental Setup}
\label{subsec:implementation}

For extracting features from the videos, we tested three different backbones, ResNet101~\cite{resnet}, I3D~\cite{i3d}, and STAM~\cite{stam}, all of them pretrained. Our approach relies on frame-level features, so for the 3D backbones, like STAM~\cite{stam} and I3D~\cite{i3d}, we densely slide a temporal window of 16 frames along each video.

Our encoder $G_e$ comprises $4$ transformer blocks, where our DTAB module is the last one, and each transformer block is composed of $h=8$ attention heads with a hidden size of $d_{model}=512$. We used the ADAM optimizer for the training schedule with a weight decay of $5 \cdot 10^{-4}$ and a learning rate of $3 \cdot 10^{-5}$ for $300$ epochs. 
We present the adversarial training, DTAB hyperparameters and sampling parameters in the supplementary material. 





\begin{sloppypar}

\subsection{Comparison with State-of-the-art Methods}
\label{subsec:comparison}

In this section, we compare our approach with different methods recently proposed in the literature.

\end{sloppypar}

\subsubsection{Results on UCF101$\leftrightarrow$HMDB51\textsubscript{full}}\label{subseq:ucfhmdb} 
As shown in~\Cref{table:ucf_hmdb_results}, we compare our results using three different backbones. The first one is ResNet101, a 2D backbone, in which our approach achieves a significant average increase of $2.4\%$, resulting in $88.2\%$. This accuracy surpasses some methods that use the I3D backbone, showing that even with a 2D backbone, our approach can extract significant spatio-temporal information. 

By analyzing methods that use the I3D backbone, our approach surpasses even works that use multi-modal data (\ie, color and motion). From single-modal data, our approach yields a significant average increase of $3.5\%$ and, compared to multi-modal methods, an average increase of $0.4\%$. 

Finally, for the STAM backbone, it can be seen in Table~\ref{table:ucf_hmdb_results} that our approach yields a significant average increase of $0.9\%$. Also, our TransferAttn model achieves a significant improvement compared to our baseline model.

\subsubsection{Results on Kinetics$\rightarrow$Gameplay}\label{subseq:kg} This section presents the results of our approach for the Kinetics$\rightarrow$ Gameplay benchmark. Since the Gameplay dataset does not give access to the raw videos, but only to frame-level features extracted from ResNet101, we were unable to provide results for methods where the backbone is not fixed in the training stage or relies on video-level features. As shown in~\Cref{table:k_g_results}, our approach achieves a significant increase of $6.5\%$ in accuracy compared with M$A^2$LT-D, a state-of-the-art method, and is better than all previous baselines. Also, it can be seen that our TransferAttn model performs significantly better than our baseline model.

\begin{table}[!htb]

\centering
\caption{Classification Accuracy on UCF101$\leftrightarrow$ HMDB51\textsubscript{full}. Multi-modal methods are represented with (C + M).}
\resizebox{8.4cm}{!}{
\begin{tabular}{l|c|ccc}
 
\hline
\hline
Method &  Backbone           & U $\rightarrow$  H & H $\rightarrow$ U & Average \\
\hline
\hline
Source Only              & \multirow{5}{*}{ResNet101}          & 73.9 & 71.7 & 72.8      \\
T$A^3$N~\cite{ta3n}     &           & 78.3 & 81.8 & 80.1      \\
M$A^2$LT-D~\cite{ma2ltd}   &           & 85.0 & 86.6 & 85.8      \\
Baseline & & 84.2& 85.4 & 84.8 \\
TransferAttn (ours)           &           & \textbf{88.1}& \textbf{88.3} & \textbf{88.2}               \\
\hline
\hline
Source Only         &      \multirow{5}{*}{TRN}          & 82.2 & 88.1 & 85.2      \\
STHC~\cite{10204931}     &                & 90.9 & 92.1 & 91.5      \\
TranSVAE~\cite{transvae}    &                 & 92.2  &   \underline{96.5}        &    \underline{94.3}           \\
Baseline          &           & 91.2& 93.6 & 92.4 \\
TransferAttn (ours)         &               & \textbf{93.5} & \textbf{97.1}  &   \textbf{95.3}    \\
\hline
\hline
Source Only         & \multirow{10}{*}{I3D}                & 80.6 & 89.3 & 85.0      \\
STCDA (C + M)~\cite{Song_2021_CVPR}   &                 & 83.1 & 92.1 &   87.7    \\
M$A^2$LT-D~\cite{ma2ltd}    &                 &   89.3   &  91.2    &   90.3        \\
C$O^2$A~\cite{co2a}      &                 & 87.8 & 95.8 & 91.8      \\
CIA (C + M)~\cite{cia}     &                 & 91.9 & 94.6 & 93.3      \\
TranSVAE~\cite{transvae}    &                 & 87.8 & \underline{99.0}          & 93.4               \\
MTRAN (C + M)~\cite{10.1145/3503161.3548009}   &                 & 92.2 & 95.3          & 93.8               \\
CleanAdapt (C + M)~\cite{cleanadapt}  &                 & 93.6 & \underline{99.3} & \underline{96.5}      \\
Baseline          &           & 90.8& 95.2 & 93.0 \\
TransferAttn (ours)          &                 & \textbf{94.4} & \textbf{99.4} & \textbf{96.9} \\
\hline
\hline
Source Only     &  \multirow{6}{*}{STAM}  & 86.9 & 93.7 & 90.3      \\
TranSVAE~\cite{transvae}     &  &   93.5   &  \underline{99.5}    &    96.5       \\
UDAVT~\cite{udavt}     &  & 92.3 & 96.8 & 94.6      \\
M$A^2$LT-D~\cite{ma2ltd}    &  &  95.3    &  \underline{99.4}    &    97.4       \\
Baseline          &           & 93.4& 98.9 & 96.1 \\
TransferAttn (ours)        &  & \textbf{97.2}          & \textbf{99.7}          & \textbf{98.5}               \\
\hline
\hline
\end{tabular}
}
\label{table:ucf_hmdb_results}
\end{table}

\begin{table}[!htb]

\centering
\caption{Classification Accuracy on Kinetics $\rightarrow$ Gameplay.}

\resizebox{5.7cm}{!}{
\begin{tabular}{l|c|c}

\hline
\hline
Method & Backbone & K $\rightarrow$ G \\
\hline
\hline
Source Only             & \multirow{6}{*}{ResNet101} & 17.6  \\
T$A^3$N~\cite{ta3n}    &          & 27.5  \\
M$A^2$LT-D~\cite{ma2ltd} &          & 31.5  \\
TranSVAE~\cite{transvae} &     &    21.9   \\
Baseline          &          & 30.4 \\
TransferAttn (ours)    &          & \textbf{37.0}  \\
\hline
\hline
\end{tabular}
}
\label{table:k_g_results}
\end{table}

\subsubsection{Results on Kinetics$\rightarrow$NEC-Drone}\label{subseq:kn} Our approach was also evaluated in the Kinetics $\rightarrow$ NEC-Drone benchmark, 
as we can see in Table~\ref{table:k_n_results}, 
achieving a significant increase of $9.5\%$ in comparison with UDAVT, 
establishing a new state-of-the-art result. 

\begin{table}[!htb]
\centering
\caption{Classification Accuracy on Kinetics $\rightarrow$ NEC-Drone.} 
\resizebox{5.7cm}{!}{
\begin{tabular}{l|c|c}

\hline
\hline
Method & Backbone              & K $\rightarrow$ N \\
\hline
\hline
Source Only              & \multirow{6}{*}{STAM} & 29.4  \\

M$A^2$LT-D~\cite{ma2ltd}  &     &   55.4    \\
TranSVAE~\cite{transvae}  &     &   55.9    \\
UDAVT~\cite{udavt}    &     & 65.3  \\
Baseline          &          & 45.5 \\
TransferAttn (ours)    &     & \textbf{74.8}  \\
\hline
\hline

\end{tabular}
}
\label{table:k_n_results}
\end{table}


\begin{table}[!htb]
\centering
\caption{Accuracy in HMDB-UCF$_{full}$ dataset integrating DTAB on state-of-the-art transformer-based video UDA architectures.}
\resizebox{7.6cm}{!}{
\begin{tabular}{l|ccc}
\hline
\hline
Method          & U $\rightarrow$ H & H $\rightarrow$ U  & Average  \\
\hline
\hline
UDAVT (Our impl.)~\cite{udavt} & 92.2 & 96.5 & 94.4 \\
UDAVT + TAM~\cite{10030518}                & 92.5 & 96.9 & 94.7  \\
UDAVT + DTAB            & \textbf{94.2} & \textbf{97.9} & \textbf{96.1}  \\
\hline
\hline
\end{tabular}
}
\label{table:udavt_with_our_transformer}
\end{table}

\subsection{Ablation Study}
\label{subsec:ablation}

This section presents an ablation study, showcasing how each design aspect of our approach impacts its performance, and shows how our DTAB can improve the results of other transformer-based methods for both action recognition (\ie, video UDA) and image classification (\ie, image UDA).

\subsubsection{Effect of DTAB components} 
\begin{figure*}[!htb]

\begin{minipage}{0.65\textwidth}
\centering
\includegraphics[height=66pt]{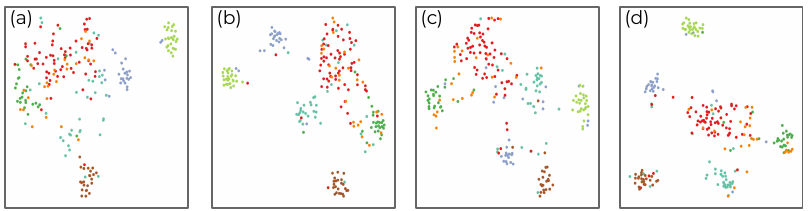}

\end{minipage}
\hfill
\begin{minipage}[b]{0.75\linewidth}

\resizebox{4.85cm}{!}{
\begin{tabular}{c|l|c}

\hline
\hline
\# & Components                             & K $\rightarrow$ N \\
\hline
\hline
a)& Baseline  & 45.5  \\
b)& + IB                      & 54.8      \\
c)& + MDTA      & 58.3  \\
\hline 
d)& DTAB (MDTA + IB)          & 74.8  \\
\hline
\hline
\end{tabular}

 }

\end{minipage}

\caption{Ablation study on Kinetics $\rightarrow$ NEC-Drone integrating each component of DTAB separately in comparison with standard transformer. \textbf{Left:} The t-SNE plots for class-wise features. \textbf{Right:} The accuracy result of each component.}
\label{fig:ablation-parts}
\end{figure*}

To understand the individual contributions from the components that integrate our DTAB module and how they improve its domain transferability, we conducted an in-depth study to evaluate how each component of DTAB impacts its performance. Figure~\ref{fig:ablation-parts} summarizes the obtained results, showing the accuracy for the Kinetics $\rightarrow$ NEC-Drone benchmark and the t-SNE visualization. 
By replacing the self-attention mechanism with our MDTA in the last transformer block, the accuracy is increased by $9.3\%$. On the other hand, using the IB mechanism in the last transformer block, even with the standard self-attention mechanism, increases the accuracy in $12.8\%$. Finally, integrating all the components, we achieve a significant increase of $29.3\%$, indicating the importance of both components in reducing the domain gap. Also, the t-SNE plots clearly show how each component impacts the clusterization of the low-dimensional features from the different classes, and it is easy to see that the combination of all the components makes features from a same class better clustered while different classes are farther away.

\subsubsection{Effects of DTAB on other methods for video UDA} 

\setcounter{table}{4}
\begin{table*}[!htb]
\centering
\caption{Classification Accuracy in Office-31 dataset integrating DTAB on state-of-the-art transformer-based image UDA architectures.}
\resizebox{12.5cm}{!}{
\begin{tabular}{l|c|ccccc|c}
\hline
\hline
Method                 & Backbone                  & A $\rightarrow$ W & D $\rightarrow$ W & A $\rightarrow$ D & D $\rightarrow$ A & W $\rightarrow$ A & Avg. \\
\hline
\hline
TVT (Our impl.)~\cite{10030518} & \multirow{2}{*}{ViT-B 16} &93.6& 98.7  & 93.7  & 79.4 & 78.7 & 88.8\\
TVT + DTAB             &         &94.7    &99.9  & 95.0 & 80.3  &    79.5   &    \textbf{89.9}    \\
\hline
CDTrans (Our impl.)~\cite{DBLP:journals/corr/abs-2109-06165} & \multirow{2}{*}{DeiT-S } & 93.5  & 98.2  & 94.0  & 77.7 & 77.0 & 88.1 \\
CDTrans + DTAB             &         &   94.7  & 98.4 & 95.6  &  77.7 &  78.1     &  \textbf{88.9}      \\
\hline
\hline
\end{tabular}
}
\label{table:tvt_with_our_transformer}
\end{table*}

Another aspect we studied is the effect of our new transformer block, DTAB, on other transformer-based methods for video UDA. For this study, we first ran UDAVT~\cite{udavt} on the UCF$\rightarrow$HMDB\textsubscript{full} benchmark and reported the results obtained by our reproduction using the author's code\footnote{\url{https://github.com/vturrisi/UDAVT}}. As a baseline, we also reported the results for the Transferability Adaption Module~(TAM), which is the attention mechanism introduced in TVT~\cite{10030518}. Due to its source-free characteristics, we were unable to evaluate DTAB on MTRAN~\cite{10.1145/3503161.3548009}.

The results in Table~\ref{table:udavt_with_our_transformer} show that adding our DTAB increased the accuracy by $1.7\%$, while TAM yielded marginal gains of less than 0.5\%. These results demonstrate that DTAB outperforms TAM in handling spatio-temporal information. Also, they suggest that our DTAB module can integrate with other transformer-based methods for video UDA, enhancing their ability to reduce the domain gap.


\subsubsection{Effects of DTAB on transformers for image UDA}

We also tested the effects of using our DTAB module with TVT~\cite{10030518} and CDTrans~\cite{DBLP:journals/corr/abs-2109-06165}, which are
transformer-based methods devised for image UDA.
For this study, experiments were conducted on the Office-31 dataset~\cite{DBLP:conf/eccv/SaenkoKFD10}, which is composed of 4,652  images from 31 classes and comprises three different domains, namely: Amazon~(A), DSRL~(D), and Webcam~(W).
The results reported in Table~\ref{table:tvt_with_our_transformer} were obtained by our reproduction using the authors' code\footnote{\url{https://github.com/uta-smile/TVT/}}$^,$\footnote{\url{https://github.com/CDTrans/CDTrans/}}. These results show a slight increase of $1.1\%$ in the accuracy of TVT~\cite{10030518}; while for CDTrans~\cite{DBLP:journals/corr/abs-2109-06165}, we observe an increase of $0.8\%$ in accuracy.

\subsubsection{Complexity Analysis} We also conduct a complexity analysis of our TransferAttn model. In this study, we compare TransferAttn and other baselines with respect to the amount of trainable parameters (\#Parameters) and floating point operations (GFLOPs). 
As shown in Table~\ref{table:complexity_analysis}, our approach has slightly more trainable parameters than TranSVAE~\cite{transvae}, however, this difference is compensated by the significant improvement in UDA results reported in Section~\ref{subsec:comparison}. Compared with other transformer-based methods, like UDAVT~\cite{udavt}, our approach has much less trainable parameters and floating point operations, in addition to a significant improvement in UDA results, indicating the efficiency of our TransferAttn model.

\setcounter{table}{5}
\begin{table}[H]
\centering
\caption{Comparison on Model Complexity}
\resizebox{6.5cm}{!}{
\begin{tabular}{l|cc}
\hline
\hline
Methods    & $\#$Parameters & GFLOPs \\
\hline
\hline
TranSVAE   & 12.7 M                 & 36.5   \\
M$A^2$LT-D & 49.7 M                 &  51.2  \\
UDAVT      & 118.9 M                & 524.7  \\
TransferAttn (ours)       & 16.3 M                 & 55.3  \\
\hline
\hline
\end{tabular}
}
\label{table:complexity_analysis}
\end{table}

\section{Conclusions}
\label{sec:conclusion}

Only a few works have exploited transformer architectures for video UDA, 
a promising strategy due to their performance in other 
tasks. Motivated by this observation, we proposed TransferAttn, a framework for video UDA and one of the few works that exploit transformer architectures to adapt cross-domain knowledge. We also propose a novel Domain Transferable-guided Attention Block~(DTAB) that employs an attention mechanism to encourage spatial-temporal transferability among video frames from different domains. 
We outperformed all other state-of-the-art methods we compared, showing the effectiveness of our TransferAttn model. Our DTAB module also demonstrated to be a promising strategy by itself and when added to other state-of-the-art UDA methods, it increased their performance.


As future work, we intend to evaluate our approach on other video UDA tasks, like video segmentation and action localization. In addition, we also plan to augment our approach to utilize multi-modal data and make it able to integrate into source-free methods. 


\paragraph{Acknowledgments}
This research was supported by São Paulo Research Foundation - FAPESP (grants \#2023/17577-0 and \#2024/04500-2) and Brazilian National Council for Scientific and Technological Development - CNPq (grants \#315220/2023-6 and \#420442/2023-5).

{\small
\bibliographystyle{ieee_fullname}
\bibliography{egbib}

\begin{thebibliography}{10}\itemsep=-1pt

\bibitem{arnab2021vivit}
Anurag Arnab, Mostafa Dehghani, Georg Heigold, Chen Sun, Mario Lu{\v{c}}i{\'c},
  and Cordelia Schmid.
\newblock Vivit: A video vision transformer.
\newblock In {\em IEEE International Conference on Computer Vision (ICCV)},
  2021.

\bibitem{carion2020end}
Nicolas Carion, Francisco Massa, Gabriel Synnaeve, Nicolas Usunier, Alexander
  Kirillov, and Sergey Zagoruyko.
\newblock End-to-end object detection with transformers.
\newblock In {\em European Conference on Computer Vision (ECCV)}, pages
  213--229. Springer, 2020.

\bibitem{k600}
Joao Carreira, Eric Noland, Andras Banki-Horvath, Chloe Hillier, and Andrew
  Zisserman.
\newblock A short note about kinetics-600.
\newblock {\em CoRR}, abs/1808.01340, 2018.

\bibitem{i3d}
Joao Carreira and Andrew Zisserman.
\newblock Quo vadis, action recognition? a new model and the kinetics dataset.
\newblock In {\em IEEE Conference on Computer Vision and Pattern Recognition
  (CVPR)}, 2017.

\bibitem{ta3n}
Min-Hung Chen, Zsolt Kira, Ghassan Alregib, Jaekwon Yoo, Ruxin Chen, and Jian
  Zheng.
\newblock Temporal attentive alignment for large-scale video domain adaptation.
\newblock In {\em IEEE International Conference on Computer Vision (ICCV)},
  pages 6320--6329, 2019.

\bibitem{ma2ltd}
Peipeng Chen, Yuan Gao, and Andy~J. Ma.
\newblock Multi-level attentive adversarial learning with temporal dilation for
  unsupervised video domain adaptation.
\newblock In {\em IEEE/CVF Winter Conference on Applications of Computer Vision
  (WACV)}, pages 776--785, 2022.

\bibitem{necdrone}
Jinwoo Choi, Gaurav Sharma, Manmohan Chandraker, and Jia-Bin Huang.
\newblock Unsupervised and semi-supervised domain adaptation for action
  recognition from drones.
\newblock In {\em IEEE/CVF Winter Conference on Applications of Computer Vision
  (WACV)}, pages 1706--1715, 2020.

\bibitem{udavt}
V. da Costa, G. Zara, P. Rota, T. Oliveira-Santos, N. Sebe, V. Murino, and E.
  Ricci.
\newblock Unsupervised domain adaptation for video transformers in action
  recognition.
\newblock In {\em IEEE International Conference on Pattern Recognition (ICPR)},
  pages 1258--1265, 2022.

\bibitem{cleanadapt}
Avijit Dasgupta, C.V. Jawahar, and Karteek Alahari.
\newblock Overcoming label noise for source-free unsupervised video domain
  adaptation.
\newblock In {\em Indian Conference on Computer Vision, Graphics and Image
  Processing (ICVGIP)}. ACM, 2023.

\bibitem{deng2023BEAR}
Andong Deng, Taojiannan Yang, and Chen Chen.
\newblock A large-scale study of spatiotemporal representation learning with a
  new benchmark on action recognition.
\newblock {\em CoRR}, abs/2303.13505, 2023.

\bibitem{donahue2015long}
Jeffrey Donahue, Lisa Anne~Hendricks, Sergio Guadarrama, Marcus Rohrbach,
  Subhashini Venugopalan, Kate Saenko, and Trevor Darrell.
\newblock Long-term recurrent convolutional networks for visual recognition and
  description.
\newblock In {\em IEEE Conference on Computer Vision and Pattern Recognition
  (CVPR)}, pages 2625--2634, 2015.

\bibitem{10.5555/3045118.3045244}
Yaroslav Ganin and Victor Lempitsky.
\newblock Unsupervised domain adaptation by backpropagation.
\newblock In {\em International Conference on Machine Learning (ICML)}, page
  1180–1189. PMLR, 2015.

\bibitem{dann}
Yaroslav Ganin, Evgeniya Ustinova, Hana Ajakan, Pascal Germain, Hugo
  Larochelle, Fran{\c{c}}ois Laviolette, Mario Marchand, and Victor Lempitsky.
\newblock {\em Domain-Adversarial Training of Neural Networks}, pages 189--209.
\newblock Springer International Publishing, 2017.

\bibitem{10.1007/978-3-319-13560-1_76}
Muhammad Ghifary, W.~Bastiaan Kleijn, and Mengjie Zhang.
\newblock Domain adaptive neural networks for object recognition.
\newblock In {\em PRICAI 2014: Trends in Artificial Intelligence}, pages
  898--904. Springer International Publishing, 2014.

\bibitem{girdhar2019video}
Rohit Girdhar, Joao Carreira, Carl Doersch, and Andrew Zisserman.
\newblock Video action transformer network.
\newblock In {\em IEEE Conference on Computer Vision and Pattern Recognition
  (CVPR)}, pages 244--253, 2019.

\bibitem{NIPS2014_5ca3e9b1}
Ian Goodfellow, Jean Pouget-Abadie, Mehdi Mirza, Bing Xu, David Warde-Farley,
  Sherjil Ozair, Aaron Courville, and Yoshua Bengio.
\newblock Generative adversarial nets.
\newblock In {\em Conference on Neural Information Processing Systems
  (NeurIPS)}, volume~27. Curran Associates, Inc., 2014.

\bibitem{resnet}
Kaiming He, Xiangyu Zhang, Shaoqing Ren, and Jian Sun.
\newblock Deep residual learning for image recognition.
\newblock In {\em IEEE Conference on Computer Vision and Pattern Recognition
  (CVPR)}, 2016.

\bibitem{CVPR_2018_Huang}
D-A. Huang, V. Ramanathan, D. Mahajan, L. Torresani, M. Paluri, L. Fei{-}Fei,
  and J.~C. Niebles.
\newblock What makes a video a video: Analyzing temporal information in video
  understanding models and datasets.
\newblock In {\em IEEE Conference on Computer Vision and Pattern Recognition
  (CVPR)}, pages 7366--7375, 2018.

\bibitem{10.1145/3503161.3548009}
Yi Huang, Xiaoshan Yang, Ji Zhang, and Changsheng Xu.
\newblock Relative alignment network for source-free multimodal video domain
  adaptation.
\newblock In {\em International Conference on Multimedia (MM)}, page
  1652–1660. ACM, 2022.

\bibitem{6165309}
Shuiwang Ji, Wei Xu, Ming Yang, and Kai Yu.
\newblock 3d convolutional neural networks for human action recognition.
\newblock {\em IEEE Transactions on Pattern Analysis and Machine Intelligence
  (TPAMI)}, 35(1):221--231, 2013.

\bibitem{k400}
Will Kay, Joao Carreira, Karen Simonyan, Brian Zhang, Chloe Hillier, Sudheendra
  Vijayanarasimhan, Fabio Viola, Tim Green, Trevor Back, Paul Natsev, Mustafa
  Suleyman, and Andrew Zisserman.
\newblock The kinetics human action video dataset.
\newblock {\em CoRR}, abs/1705.06950, 2017.

\bibitem{ke2017new}
Qiuhong Ke, Mohammed Bennamoun, Senjian An, Ferdous Sohel, and Farid Boussaid.
\newblock A new representation of skeleton sequences for 3d action recognition.
\newblock In {\em IEEE Conference on Computer Vision and Pattern Recognition
  (CVPR)}, pages 3288--3297, 2017.

\bibitem{2020arXiv200308264K}
Donghyun {Kim}, Kuniaki {Saito}, Tae-Hyun {Oh}, Bryan~A. {Plummer}, Stan
  {Sclaroff}, and Kate {Saenko}.
\newblock {Cross-domain Self-supervised Learning for Domain Adaptation with Few
  Source Labels}.
\newblock {\em CoRR}, abs/2003.08264, 2020.

\bibitem{kim2024adversarial}
Kiyoon Kim, Shreyank~N Gowda, Panagiotis Eustratiadis, Antreas Antoniou, and
  Robert~B Fisher.
\newblock Adversarial augmentation training makes action recognition models
  more robust to realistic video distribution shifts.
\newblock {\em CoRR}, abs/2401.11406, 2024.

\bibitem{kong2022human}
Yu Kong and Yun Fu.
\newblock Human action recognition and prediction: A survey.
\newblock {\em International Journal of Computer Vision (IJCV)},
  130(5):1366--1401, 2022.

\bibitem{hmdb51}
H. Kuehne, H. Jhuang, E. Garrote, T. Poggio, and T. Serre.
\newblock Hmdb: A large video database for human motion recognition.
\newblock In {\em IEEE International Conference on Computer Vision (ICCV)},
  pages 2556--2563, 2011.

\bibitem{lai2024empowering}
Zhengfeng Lai, Haoping Bai, Haotian Zhang, Xianzhi Du, Jiulong Shan, Yinfei
  Yang, Chen-Nee Chuah, and Meng Cao.
\newblock Empowering unsupervised domain adaptation with large-scale
  pre-trained vision-language models.
\newblock In {\em IEEE/CVF Winter Conference on Applications of Computer Vision
  (WACV)}, pages 2691--2701, 2024.

\bibitem{10204931}
K. Li, D. Patel, E. Kruus, and M. Min.
\newblock Source-free video domain adaptation with spatial-temporal-historical
  consistency learning.
\newblock In {\em IEEE Conference on Computer Vision and Pattern Recognition
  (CVPR)}, pages 14643--14652, 2023.

\bibitem{liu2016spatio}
Jun Liu, Amir Shahroudy, Dong Xu, and Gang Wang.
\newblock Spatio-temporal lstm with trust gates for 3d human action
  recognition.
\newblock In {\em European Conference on Computer Vision (ECCV)}, pages
  816--833. Springer, 2016.

\bibitem{liu2021swin}
Ze Liu, Yutong Lin, Yue Cao, Han Hu, Yixuan Wei, Zheng Zhang, Stephen Lin, and
  Baining Guo.
\newblock Swin transformer: Hierarchical vision transformer using shifted
  windows.
\newblock In {\em IEEE International Conference on Computer Vision (ICCV)},
  pages 10012--10022, 2021.

\bibitem{10.5555/3305890.3305909}
Mingsheng Long, Han Zhu, Jianmin Wang, and Michael~I. Jordan.
\newblock Deep transfer learning with joint adaptation networks.
\newblock In {\em International Conference on Machine Learning (ICML)}, page
  2208–2217. PMRL, 2017.

\bibitem{mm-sada}
Jonathan Munro and Dima Damen.
\newblock {M}ulti-modal {D}omain {A}daptation for {F}ine-grained {A}ction
  {R}ecognition.
\newblock In {\em IEEE Conference on Computer Vision and Pattern Recognition
  (CVPR)}, 2020.

\bibitem{2020arXiv200610297P}
Changhwa {Park}, Jonghyun {Lee}, Jaeyoon {Yoo}, Minhoe {Hur}, and Sungroh
  {Yoon}.
\newblock {Joint Contrastive Learning for Unsupervised Domain Adaptation}.
\newblock {\em CoRR}, abs/2006.10297, 2020.

\bibitem{DBLP:conf/eccv/SaenkoKFD10}
Kate Saenko, Brian Kulis, Mario Fritz, and Trevor Darrell.
\newblock Adapting visual category models to new domains.
\newblock In Kostas Daniilidis, Petros Maragos, and Nikos Paragios, editors,
  {\em European Conference on Computer Vision (ECCV)}, pages 213--226.
  Springer, 2010.

\bibitem{shahroudy2016ntu}
Amir Shahroudy, Jun Liu, Tian-Tsong Ng, and Gang Wang.
\newblock Ntu rgb+ d: A large scale dataset for 3d human activity analysis.
\newblock In {\em IEEE Conference on Computer Vision and Pattern Recognition
  (CVPR)}, pages 1010--1019, 2016.

\bibitem{stam}
Gilad Sharir, Asaf Noy, and Lihi Zelnik-Manor.
\newblock An image is worth 16x16 words, what is a video worth?
\newblock {\em CoRR}, abs/2103.13915, 2021.

\bibitem{si2019attention}
Chenyang Si, Wentao Chen, Wei Wang, Liang Wang, and Tieniu Tan.
\newblock An attention enhanced graph convolutional lstm network for
  skeleton-based action recognition.
\newblock In {\em IEEE Conference on Computer Vision and Pattern Recognition
  (CVPR)}, pages 1227--1236, 2019.

\bibitem{Song_2021_CVPR}
Xiaolin Song, Sicheng Zhao, Jingyu Yang, Huanjing Yue, Pengfei Xu, Runbo Hu,
  and Hua Chai.
\newblock Spatio-temporal contrastive domain adaptation for action recognition.
\newblock In {\em IEEE Conference on Computer Vision and Pattern Recognition
  (CVPR)}, pages 9787--9795, 2021.

\bibitem{ucf101}
Khurram Soomro, Amir~Roshan Zamir, and Mubarak Shah.
\newblock {UCF101:} {A} dataset of 101 human actions classes from videos in the
  wild.
\newblock {\em CoRR}, abs/1212.0402, 2012.

\bibitem{7410867}
D. Tran, L. Bourdev, R. Fergus, L. Torresani, and M. Paluri.
\newblock Learning spatiotemporal features with 3d convolutional networks.
\newblock In {\em IEEE International Conference on Computer Vision (ICCV)},
  pages 4489--4497, 2015.

\bibitem{co2a}
Victor~G. Turrisi~da Costa, Giacomo Zara, Paolo Rota, Thiago Oliveira-Santos,
  Nicu Sebe, Vittorio Murino, and Elisa Ricci.
\newblock Dual-head contrastive domain adaptation for video action recognition.
\newblock In {\em IEEE/CVF Winter Conference on Applications of Computer Vision
  (WACV)}, pages 2234--2243, 2022.

\bibitem{8099799}
Eric Tzeng, Judy Hoffman, Kate Saenko, and Trevor Darrell.
\newblock Adversarial discriminative domain adaptation.
\newblock In {\em IEEE Conference on Computer Vision and Pattern Recognition
  (CVPR)}, pages 2962--2971, 2017.

\bibitem{7508942}
Keze Wang, Dongyu Zhang, Ya Li, Ruimao Zhang, and Liang Lin.
\newblock Cost-effective active learning for deep image classification.
\newblock {\em IEEE Transactions on Circuits and Systems for Video Technology
  (TCSVT)}, 27(12):2591--2600, 2017.

\bibitem{wang2021tdn}
Limin Wang, Zhan Tong, Bin Ji, and Gangshan Wu.
\newblock Tdn: Temporal difference networks for efficient action recognition.
\newblock In {\em IEEE Conference on Computer Vision and Pattern Recognition
  (CVPR)}, pages 1895--1904, 2021.

\bibitem{wang2018temporal}
Limin Wang, Yuanjun Xiong, Zhe Wang, Yu Qiao, Dahua Lin, Xiaoou Tang, and Luc
  Van~Gool.
\newblock Temporal segment networks for action recognition in videos.
\newblock {\em IEEE Transactions on Pattern Analysis and Machine Intelligence
  (TPAMI)}, 41(11):2740--2755, 2018.

\bibitem{transvae}
Pengfei Wei, Lingdong Kong, Xinghua Qu, Yi Ren, zhiqiang xu, Jing Jiang, and
  Xiang Yin.
\newblock Unsupervised video domain adaptation for action recognition: A
  disentanglement perspective.
\newblock In {\em Conference on Neural Information Processing Systems
  (NeurIPS)}. Curran Associates, Inc., 2023.

\bibitem{wu2015modeling}
Zuxuan Wu, Xi Wang, Yu-Gang Jiang, Hao Ye, and Xiangyang Xue.
\newblock Modeling spatial-temporal clues in a hybrid deep learning framework
  for video classification.
\newblock In {\em International Conference on Multimedia (MM)}, pages 461--470.
  ACM, 2015.

\bibitem{DBLP:journals/corr/abs-2109-06165}
Tongkun Xu, Weihua Chen, Pichao Wang, Fan Wang, Hao Li, and Rong Jin.
\newblock Cdtrans: Cross-domain transformer for unsupervised domain adaptation.
\newblock {\em CoRR}, abs/2109.06165, 2021.

\bibitem{yan2018spatial}
Sijie Yan, Yuanjun Xiong, and Dahua Lin.
\newblock Spatial temporal graph convolutional networks for skeleton-based
  action recognition.
\newblock In {\em AAAI Conference on Artificial Intelligence}, volume~32, 2018.

\bibitem{10030518}
Jinyu Yang, Jingjing Liu, Ning Xu, and Junzhou Huang.
\newblock Tvt: Transferable vision transformer for unsupervised domain
  adaptation.
\newblock In {\em IEEE/CVF Winter Conference on Applications of Computer Vision
  (WACV)}, pages 520--530, 2023.

\bibitem{cia}
Lijin Yang, Yifei Huang, Yusuke Sugano, and Yoichi Sato.
\newblock Interact before align: Leveraging cross-modal knowledge for domain
  adaptive action recognition.
\newblock In {\em IEEE Conference on Computer Vision and Pattern Recognition
  (CVPR)}, pages 14702--14712, 2022.

\bibitem{mix-dann}
Yuehao Yin, Bin Zhu, Jingjing Chen, Lechao Cheng, and Yu-Gang Jiang.
\newblock Mix-dann and dynamic-modal-distillation for video domain adaptation.
\newblock In {\em International Conference on Multimedia (MM)}, page
  3224–3233. ACM, 2022.

\bibitem{yue2015beyond}
Joe Yue-Hei~Ng, Matthew Hausknecht, Sudheendra Vijayanarasimhan, Oriol Vinyals,
  Rajat Monga, and George Toderici.
\newblock Beyond short snippets: Deep networks for video classification.
\newblock In {\em IEEE Conference on Computer Vision and Pattern Recognition
  (CVPR)}, pages 4694--4702, 2015.

\bibitem{zhu2016co}
Wentao Zhu, Cuiling Lan, Junliang Xing, Wenjun Zeng, Yanghao Li, Li Shen, and
  Xiaohui Xie.
\newblock Co-occurrence feature learning for skeleton based action recognition
  using regularized deep lstm networks.
\newblock In {\em AAAI Conference on Artificial Intelligence}, volume~30, 2016.

\end{thebibliography}


\begin{thebibliography}{1}\itemsep=-1pt

\bibitem{ta3n}
Min-Hung Chen, Zsolt Kira, Ghassan Alregib, Jaekwon Yoo, Ruxin Chen, and Jian
  Zheng.
\newblock Temporal attentive alignment for large-scale video domain adaptation.
\newblock In {\em IEEE International Conference on Computer Vision (ICCV)},
  pages 6320--6329, 2019.

\bibitem{necdrone}
Jinwoo Choi, Gaurav Sharma, Manmohan Chandraker, and Jia-Bin Huang.
\newblock Unsupervised and semi-supervised domain adaptation for action
  recognition from drones.
\newblock In {\em IEEE/CVF Winter Conference on Applications of Computer Vision
  (WACV)}, pages 1706--1715, 2020.

\bibitem{udavt}
V. da Costa, G. Zara, P. Rota, T. Oliveira-Santos, N. Sebe, V. Murino, and E.
  Ricci.
\newblock Unsupervised domain adaptation for video transformers in action
  recognition.
\newblock In {\em IEEE International Conference on Pattern Recognition (ICPR)},
  pages 1258--1265, 2022.

\end{thebibliography}
}

\end{document}


\title{Supplementary Materials for paper Transferable-guided Attention Is All You Need for Video Domain Adaptation}  

\maketitle
\thispagestyle{empty}
\appendix

\section{Experimental Details}
\label{sec:intro}

The adversarial training needs the hyperparameters $\lambda$ to control the strength of the Gradient Reversal Layer~(GRL) on the adaptation head. Our DTAB module needs the hyperparameters $\mathcal{Q}$ and $\alpha$, where the first controls the queue size and the other controls the weight of the IB loss~\cite{udavt} and, in the GRL from DTAB, we fixed the weight $\lambda_{DTAB}$ as 1 for every adaptation task. Also, we must define the batch size and the $k$ sampling frames related to the training schedule.

For the \textbf{UCF} $\rightarrow$ \textbf{HMDB}~\cite{ta3n} benchmark, we used a batch size of $32$ and a sample of $k=53$ frames. Due to its smaller size, we used a queue size $\mathcal{Q}$ of $1024$. Also, we used the IB loss of $\alpha = 0.001$ and adversarial loss of $\lambda = 1$. In the \textbf{HMDB} $\rightarrow$ \textbf{UCF} benchmark, the only change is in the adversarial loss of $\lambda = 0.5$ to make the training more stable.

For the \textbf{Kinetics} $\rightarrow$ \textbf{Gameplay}~\cite{ta3n} benchmark, we used a batch size of $64$ and a sample of $k=23$ frames. In this adaptation task, we reduced the queue size $\mathcal{Q}$ to $512$, used an IB loss of $\alpha = 0.001$, and a minor adversarial loss of $\lambda = 0.05$, making the training more stable.

In the \textbf{Kinetics} $\rightarrow$ \textbf{NEC-Drone}~\cite{necdrone} benchmark, we used a batch size of $64$, a sample of $k=53$ frames, a queue size $\mathcal{Q}$ of $512$, an IB loss of $\alpha = 0.025$, and an adversarial loss of $\lambda = 0.5$.

\section{More Ablation Studies}
\label{sec:moreexp}

This section reports the extra ablation studies conducted with our TransferAttn framework.

\subsection{Effect of the DTAB position} 

To study the impact of the position of the DTAB module, we experimented by first changing all transformer blocks to DTAB, then changing only the first and last ones, and finally placing them in odd and even positions. The results in Table~\ref{table:k_n_results_ablation_positions} show that our DTAB works better when used in the place of the last transformer block, where the patch features are more fine-grained than the others.

\begin{table}[h!]
\centering
\caption{Ablation study on Kinetics $\rightarrow$ NEC-Drone integrating the DTAB in different encoder positions.}
\begin{tabular}{l|c|c}
\hline
\hline
DTAB Position  & Backbone              & K $\rightarrow$ N \\
\hline
\hline
All Blocks     & \multirow{5}{*}{STAM} &  36.0     \\
First Only     &     &   38.2    \\
Even Positions &     &   54.0    \\
Odd Positions  &     &    65.4   \\
Last Only      &     &  \textbf{74.8}    \\
\hline
\hline
\end{tabular}
\label{table:k_n_results_ablation_positions}
\end{table}

\subsection{Effect of the Fixed Classifier}

One of the hypotheses we introduce in this paper is the use of a classifier with fixed random weights. This approach is motivated by the idea that fixing the classification boundaries forces the encoder $G_e$ to learn a feature space that is more generalizable across domains, avoiding the classification head $G_C$ to overfit on the source domain data. To study the impact of the fixed random classifier, we propose an ablation study to evaluate both our baseline and TransferAttn models with both learnable and fixed classifiers.

In Table~\ref{table:k_n_ablation_classifier}, we present the accuracy results on the Kinetics $\rightarrow$ NEC-Drone benchmark, using the STAM backbone. As we can see, in both models, the use of a fixed  random classifier yields an improvement in the final result, demonstrating that fixing the classification boundaries makes the encoder $G_e$ learn more robust features for UDA.

\begin{table}[h!]
\centering
\caption{Ablation study on Kinetics $\rightarrow$ NEC-Drone using learned and fixed classifier.}
\begin{tabular}{l|l|c}
\hline
\hline
Method  & Classifier            & K $\rightarrow$ N \\
\hline
\hline
\multirow{2}{*}{Baseline}  & Learnable  &  41.7    \\
  & Fixed  &  \textbf{45.5}     \\
  \hline 
\multirow{2}{*}{TransferAttn}  & Learnable  &  69.2    \\
  & Fixed  &   \textbf{74.8}     \\

\hline
\hline
\end{tabular}
\label{table:k_n_ablation_classifier}
\end{table}

{\small
\bibliographystyle{ieee_fullname}
\bibliography{egbib}
}